%% file: paper.tex
\documentclass{article}
\usepackage{spconf,amsmath,graphicx}

\usepackage[hidelinks]{hyperref}

\makeatletter
\def\bstctlcite{\@ifnextchar[{\@bstctlcite}{\@bstctlcite[@auxout]}}
\def\@bstctlcite[#1]#2{\@bsphack
  \@for\@citeb:=#2\do{%
    \edef\@citeb{\expandafter\@firstofone\@citeb}%
    \if@filesw\immediate\write\csname #1\endcsname{\string\citation{\@citeb}}\fi}%
  \@esphack}
\makeatother 

\usepackage{enumitem}
\setlist{nosep, leftmargin=14pt}

\usepackage{booktabs}
\usepackage{flushend}

\newcommand{\myref}[2]{\hyperref[#2]{#1~\ref*{#2}}}

\title{Beyond Occlusion: In Search for Near Real-Time Explainability of CNN-Based Prostate Cancer Classification}
\name{Martin Krebs \qquad Jan Obdržálek \qquad Vít Musil \qquad Tomáš Brázdil}

\address{Faculty of Informatics, Masaryk University, Brno, Czech Republic}

\begin{document}
\bstctlcite{IEEEexample:BSTcontrol}
\maketitle
\begin{abstract}
Deep neural networks are starting to show their worth in critical applications such as assisted cancer diagnosis.
However, for their outputs to get accepted in practice, the results they provide should be explainable in a way easily understood by pathologists.
A well-known and widely used explanation technique is occlusion, which, however, can take a long time to compute, thus slowing the development and interaction with pathologists.
In this work, we set out to find a faster replacement for occlusion in a successful system for detecting prostate cancer.
Since there is no established framework for comparing the performance of various explanation methods, we first identified suitable comparison criteria and selected corresponding metrics.
Based on the results, we were able to choose a different explanation method, which cut the previously required explanation time at least by a factor of 10, without any negative impact on the quality of outputs.
This speedup enables rapid iteration in model development and debugging and brings us closer to adopting AI-assisted prostate cancer detection in clinical settings.
We propose that our approach to finding the replacement for occlusion can be used to evaluate candidate methods in other related applications.
\end{abstract}
\section{Introduction}
\label{sec:introduction}

Prostate cancer is the most common oncological disease among men in the Czech Republic.
Given its high prevalence and potential for aggressiveness, early detection is crucial to improve outcomes and survival rates \cite{early-detection}.
Digital pathology, the contemporary approach to prostate cancer detection, involves analyzing digitized copies of prostate tissue samples, called Whole Slide Images (WSI's).
WSIs are displayed in a dedicated computer browser, enabling quick navigation and introspection of the digitized tissue.
Detection of prostate cancer in these digital samples relies on identifying a set of relatively well-established morphological patterns that visually differ from the surrounding healthy tissue \cite{gleason-patterns}.

In~\cite{gallo}, the authors developed and trained a convolutional neural network (CNN) model to detect prostate cancer.
However, the main contribution of~\cite{gallo} is that they showed how to develop a model that not only correctly identifies prostate cancer but also demonstrably does so by utilizing the very patterns used for this purpose by trained pathologists.
To achieve this goal, they relied on an explainability technique known as occlusion, which works by replacing the most salient regions of the input image and observing the effect on predictions produced by the network.

While occlusion works well in this application, it is too slow for near-real-time assistance or explaining large volumes of data, as it needs to execute multiple forward passes of the network.
Can this be improved by using a different, faster explainability method? While there are many faster methods, including, but not limited to, various CAM-based methods \cite{CAM, GradCAM, GradCAMplusplus, HiResCAM}, not all are suitable for our use case.
We therefore need to compare explainability methods according to their speed, accuracy, and understandability of outputs produced.

The problem with finding a suitable method is that comparing various explainability approaches is an inherently hard task.
And while many comparison metrics have been developed, e.g.~\cite{ROAD, ROAR, ehr, weighting-game}, they usually only address one particular aspect, e.g.
accuracy.
We therefore need a more holistic approach to comparing method performance.

For our prostate cancer setting of \cite{gallo}, we were able to identify several comparison criteria and select the corresponding metrics.
We then chose four candidate well-established,  single-pass methods -- CAM \cite{CAM}, GradCAM++ \cite{GradCAMplusplus}, HiResCAM \cite{HiResCAM}, and Composite-LRP \cite{LRP} (some of which leverage the architecture of our network) -- and evaluated them according to our criteria.
We found out that some of these methods generally perform at least as well as occlusion in our application.
The finally chosen methods were then brought into our development chain, reducing the turnaround time between consecutive evaluations by pathologist, leading to faster model development.
We believe that our approach can be, possibly after minor domain-necessitated changes, also used to evaluate candidate explanation methods for other applications of CNNs in digital pathology and beyond.

\section{Materials \& Methods}
\label{sec:methods}

\subsection{Data}

We use the test partition of the hematoxylin/eosin-stained WSI dataset from \cite{gallo}, comprising $87$ biopsies from $10$ patients.
Each WSI holds $3$ to $5$ biopsies and is stored as an uncompressed PNG of size $105,185 \times 221,772$ pixels.
While each case is from a patient with a positive diagnosis, some of the biopsies are non-cancerous, yielding the positive/negative ratio of WSIs of $37/50$.
The distribution of WHO grade groups from $1$ to $5$ across patients is 43, 32, 17, 10, 12.
We have a pixel-level precise annotation for each positive WSI in the form of a polygon encapsulating the respective cancerous tissue.
The WSIs were, following the standard approach, sliced into overlapping patches of $512 \times 512$ pixels.
We overlap the patches by $256$ pixels to counter the chance of severing important morphological patterns.
We process the WSIs at level 1, which corresponds to 10\texttimes\ magnification and resolution $0.344$~µm per pixel.

\subsection{Model}

Our model (see~\cite{gallo} for a detailed description) is a binary classifier utilizing a VGG-16 backbone, followed by a global max pool layer and a single fully connected layer.
It is trained to determine whether a patch's central $256 \times 256$ pixels area contains cancerous tissue.
The label for a patch is based on whether the central area overlaps with the annotation.
Despite the model's relatively small size and simplicity, it achieves respectable performance on the test set, with 100\% slide-level accuracy.

\subsection{Explainability Methods}

The method currently in place is occlusion.
It works by systematically covering regions in the input and observing changes in model's confidence.
While this approach produces comprehensible and faithful explanations, it is computationally slow.
To explain a single tile, we need multiple forward passes.
As a result, computing one saliency mask for a WSI takes anywhere from 30 to 90 minutes, depending on the number of tissue-containing tiles.
To replace occlusion, we consider four methods: CAM, GradCAM++, HiResCAM, and Composite-LRP.
We have chosen these methods according to the following criteria: a) methods which have a theoretical potential to be fast in our case, and b) methods which were not previously tested in~\cite{gallo}.

\textbf{CAM}
\cite{CAM} is a model-specific method that works only for networks that contain a global average (or maximum) pooling layer between the last convolutional layer and the final fully connected layer, a restriction which our model satisfies.
CAM (as all CAM-based methods) utilizes spatial information in the convolutional layers.
We use the implementation from the \texttt{torch-cam}\footnote{\url{https://github.com/frgfm/torch-cam/}} library.

\textbf{GradCAM++}
\cite{GradCAMplusplus} is gradient-based model-agnostic method, which addresses some shortcomings of the well-known gradient-based GradCAM~\cite{GradCAM} method.
The reason we include GradCAM++ is that due to its design, the resulting saliency maps significantly differ from those produced by CAM.
We use the \texttt{python-grad-cam}\footnote{\label{lib:pgc}\url{https://github.com/jacobgil/pytorch-grad-cam}} library implementation.

\textbf{HiResCAM}
\cite{HiResCAM} is another of the gradient-based methods, for which it has been proven in~\cite{HiResCAM} (under some assumptions) that it is guaranteed to highlight all parts of image which increase the class score.
However, these theoretical guarantees do not directly apply to our model because of the presence of the global max pooling layer.
We use the implementation from the \texttt{python-grad-cam}\footnotemark[2] library.

\textbf{Composite-LRP}
Layer-Wise Relevant Propagation~\cite{LRP} is a method which uses various rules to propagate the score of the output layer to individual pixels.
We use the LRP method implementation from the~\texttt{captum}\footnote{\url{https://github.com/pytorch/captum}} library, with the following modification: To obtain visually sharper results, we replace the LRP-$\gamma$ rule with the LRP-$\alpha,\beta$ rule from~\cite{LRPalphabeta}.

\subsection{Metrics for comparing methods}

As we have mentioned earlier, comparing explainability methods is hard, and there is no consensus on which metrics should be used to evaluate the many approaches to explainability.
Here we suggest four \emph{quantitative} metrics, which form a general framework for evaluating explainability methods and show how these metrics can be applied to a model like ours.
We first measure the amount of computational resources needed, while the following three metrics are intended to assess the quality of explanations.

\textbf{Computational performance}
The high amounts of time needed to produce good explanations have been the original motivation for this research.
Methods that take long time to produce results can significantly impact adoption by pathologists who work in high-pressure environment.
We find it helpful to measure both the time needed for an individual slide and for the whole slide (WSI).
The other aspect to watch for is GPU memory usage, as high GPU utilization can limit the number of processes that can run in parallel, which in turn increases the deployment costs.

To evaluate our methods, we let each method to explain a single WSI of $1,499$ tiles.
We run \texttt{nvidia-smi} process with 500~ms sampling frequency to monitor the GPU usage.
All experiments were performed on an 8-core AMD EPYC™ 9454-based machine with 16 GB of RAM and a single NVIDIA A40 GPU card with 48 GB memory.

\textbf{Faithfulness}
In an ideal world, an explainability method highlights only those regions which the model considers important for its decision.
A common approach is perturbing the original image by replacing examined regions with a fixed value and reevaluating the model on the perturbed input.
However, approaches based on perturbation are often prone to introducing unintentional artifacts.
Therefore, alternative metrics have been produced, such as Remove and Retrain (ROAR)~\cite{ROAR}, which involves zeroing out the identified features and retraining the model from scratch.
However, ROAR can also suffer from information leakage, as the shape of the uniformly-valued removed region influences model training.
In this study, we therefore use a method called Remove and Debias (ROAD)~\cite{ROAD}, which reduces the risk of information leakage.
As in the original paper~\cite{ROAD}, we evaluate our methods at 10, 20, 30, 40, and 50 percent of the most salient perturbed pixels.

\textbf{Localization}
While faithfulness tells us how well an explainability method associates the important features with model decisions, this is typically not enough for clinical practice.
What a pathologist expects is that those features resemble the structure the pathologist is looking for -- i.e., adenocarcinoma in our case.
We therefore want to measure how closely the output of a method matches pathologist's annotations.
To this end, the original paper~\cite{gallo} uses a method called Effective Heat Ratio~\cite{ehr}, which relies on bounding boxes in ground truth data locating the objects of interest.
Arguably in our case, EHR is not the most appropriate method, as it can discriminate methods which point to correct markers but do not mark the healthy surrounding tissue included in the annotation~\cite{gallo, annotation-agreement}.
Instead, we go with a simpler metric called Weighting Game~\cite{weighting-game}, which builds upon the well-established Pointing Game metric~\cite{pointing-game}.
This metric calculates the ratio of saliency mass within the bounding box with respect to saliency mass for the whole image.

\textbf{Usefulness}
To be useful in practice, domain experts must be able to easily understand the explanations produced.
Since this is of course a highly subjective area, can we actually quantify this with any reasonable accuracy?  Fortunately, here we can exploit the fact shown in~\cite{gallo} that occlusion generates semantically correct saliency maps.
We can therefore again employ the aforementioned Weighting Game metric, this time measuring how well the produced saliency maps resemble the maps produced by occlusion.

\section{Results}
\label{sec:results}

\textbf{Computation performance}
The results are summarized in Table~\ref{tab:comp-time}.
There is no surprise there, as all the evaluated methods require just a single pass to compute saliency maps, unlike occlusion, which is perturbance based and thus requires repeated evaluation of the input.
Also, note the penalty incurred for more sophisticated variants of the basic CAM method.

\begin{table}[h!]
\centering
\begin{tabular}{@{} l @{\ }c r r @{}}\toprule
Method & tile time & \multicolumn{1}{c}{slide time} & \multicolumn{1}{c}{\ GPU use}\\
\midrule
CAM                 & $0.04 \pm 0.04$ s & $0.99$ m  & $1,084$ MB\\
GradCAM++           & $0.08 \pm 0.00$ s & $2.19$ m  & $1,578$ MB\\
HiResCAM            & $0.08 \pm 0.00$ s & $2.20$ m  & $1,578$ MB\\
Composite-LRP       & $0.11 \pm 0.00$ s & $2.98$ m  & $2,476$ MB\\
\textbf{Occlusion}  & $1.06 \pm 0.00$ s & $26.87$ m & $36,308$ MB\\
\bottomrule
\end{tabular}
\caption{Time efficiency}
\label{tab:comp-time}
\end{table}

GPU utilization follows a similar pattern to time efficiency.
The higher memory usage of Composite-LRP is caused by the method running a complete backward pass equivalent and needs to remember forward pass activations to compute the relevance scores.

\begin{figure}[hb]
    \centering
    \includegraphics[width=\linewidth]{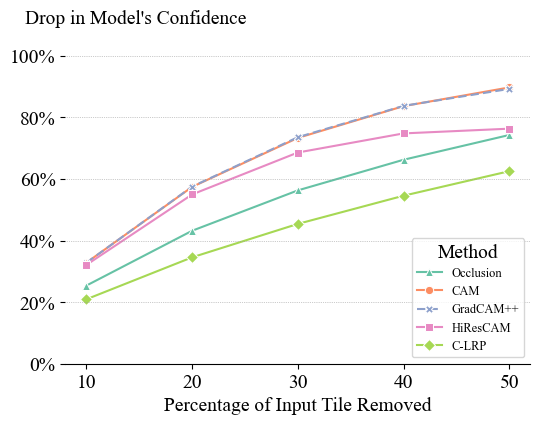}
    \caption{Faithfulness according to the ROAD method.}
    \label{fig:res}
\end{figure}

\textbf{Faithfulness}
The results are summarized in Fig.~\ref{fig:res}.
For a good explainability method, we want the score to drop markedly when even a small percentage of features (based on the saliency maps) is removed.
As we can see, occlusion actually outperforms Composite-LRP, in agreement with~\cite{gallo}.
However all CAM-based methods perform significantly better than either of them.
And while HiResCAM is for larger percentages outperformed by both CAM and GradCAM++, its curve also clearly exhibits the desired behavior -- that the curve should flatten when a significant fraction of the input is removed.

\begin{figure}
    \centering
    \includegraphics[width=\linewidth]{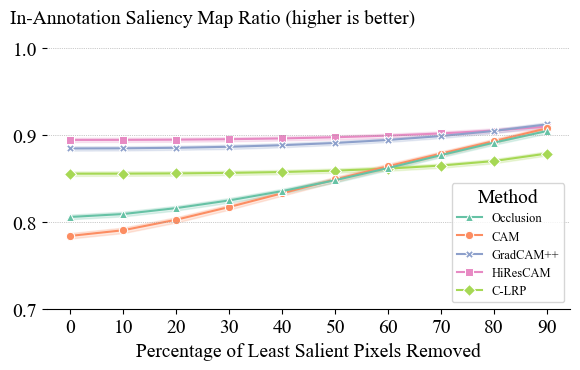}
    \caption{Agreement with pathologist annotations.}
    \label{fig:wg_annot}
\end{figure}

\textbf{Localization}
In~Fig.~\ref{fig:wg_annot}, we see how much the saliency map produced by each method agrees with pathologists annotation.
Both gradient-based CAM methods clearly outperform the rest, especially in the low saliency areas.
However, the size of the difference between CAM and GradCAM++ has been quite puzzling.
A visual inspection revealed that CAM generally covers a larger share of tile area and therefore catches up with GradCAM++ when only the most salient regions are considered.
While Composite-LRP starts with a better score than occlusion, its performance does not improve much for higher percentages of removed saliency.
A probable explanation is that this method focuses on features that do not fall within the annotation mask.

\begin{figure}[htb]
    \centering
    \includegraphics[width=\linewidth]{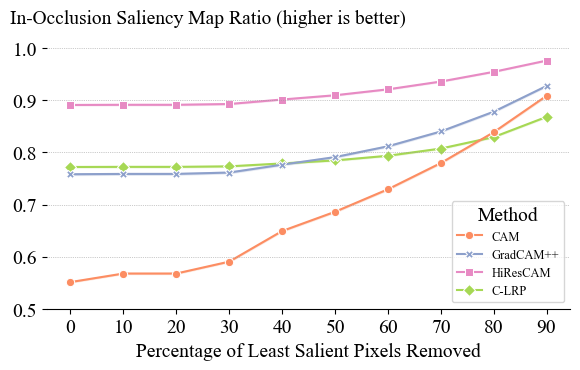}
    \caption{Agreement with
      occlusion-produced maps.}
    \label{fig:wg_occ}
\end{figure}

\textbf{Usefulness}
As can be seen in~Fig.~\ref{fig:wg_occ}, HiResCAM consistently achieves high ratios, meaning the explanations produced agree with occlusion and, by~\cite{gallo}, work well for pathologists.
This measure also, perhaps surprisingly, significantly differentiates CAM from GradCAM++, which performs significantly better in this aspect and tends to better resemble the occlusion-based annotation.
The even performance of Composite-LRP lends more weight to the argument that it assigns saliency to regions not considered important to other methods.

\textbf{Domain expert assessment}
To complement the quantitative findings above, we have chosen two of the most promising methods, CAM and HiResCAM, for a qualitative assessment by our local pathologist, Dr.
Nenutil, who has significant experience in this domain.
The decision of choosing CAM instead of GradCAM++ was based on the fact that it produces similar results as GradCAM++, while being faster.
Also, unlike GradCAM++, its saliency maps also contain negative regions, pointing to the areas which the model perceives as influencing the score against cancer -- similar to occlusion.

\begin{figure}[htb]
    \centering
    \includegraphics[width=\linewidth]{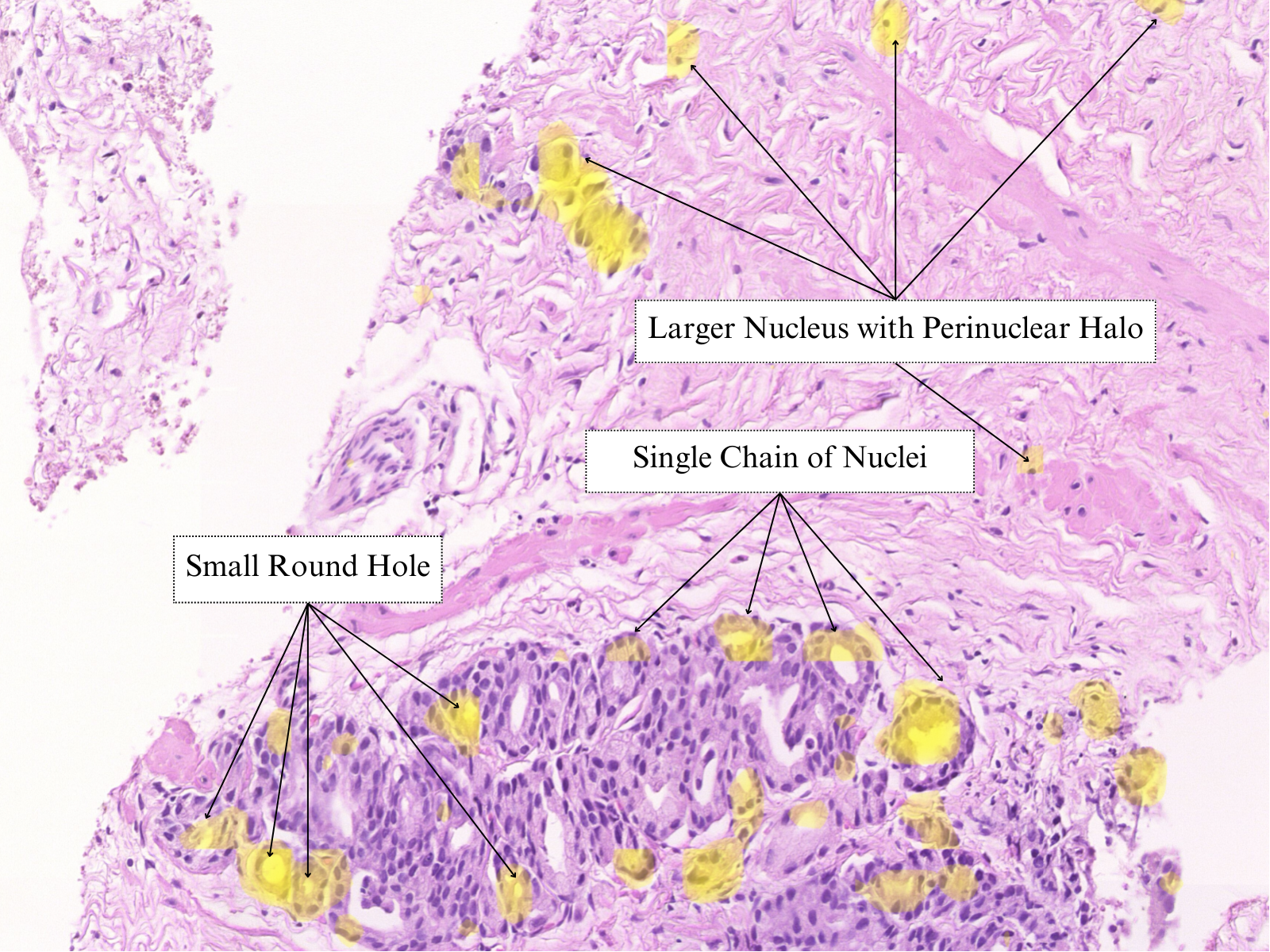}
    \caption{Annotated saliency map produced by HiResCAM.}
    \label{fig:hires_output}
\end{figure}

We have given both the 87 WSIs and the maps produced by HiResCAM and CAM to the expert to assess whether the maps produced by these two methods could be used instead of the occlusion maps.
(We intentionally withheld the occlusion maps to avoid influencing the expert's opinion.) According to him, both methods successfully detect the pro-cancer patterns from~\cite{gallo}, and CAM additionally successfully detects the non-cancer tissue patterns.
His specific observations were that ``CAM is more sensitive to `High nuclear density' positive pattern'' and ``HiResCAM is more sensitive to `Single chain of nuclei' and `Larger nucleus with perinuclear halo' patterns, which results in much more false positive labels''.
An example of an annotated HiResCAM output is in Fig.~\ref{fig:hires_output}.

\section{Conclusion}
\label{sec:conclusion}

Our goal was to find an explainability method that could replace occlusion in the setting of~\cite{gallo}.
The requirements were significantly smaller running times, while preserving or even improving accuracy.
To compare the candidate explainability methods, we have identified several quantitative metrics as a benchmark to steer us toward the ideal candidate.
All CAM-based methods performed better than both occlusion and Composite-LRP, with the gradient-based method outperforming basic CAM.
Combining the results of our benchmark with the domain expert's assessment, HiResCAM turned out to be the best method.
By incorporating our own implementation of HiResCAM in our toolkit, we were able to decrease the time needed to compute the saliency maps for all of our 87 WSIs to a mere 2 hours, from the 3.3 days needed before.
We believe that the approach we used can serve as a general template for benchmarking explainability methods in settings similar to ours.

\section{Compliance with ethical standards}
\label{sec:ethics}

The project was approved by the Ethical Committee of Masaryk Memorial Cancer Institute, No. MOU 385 920.

\section{Acknowledgments}
\label{sec:acknowledgments}

We would like to thank Dr.\ Rudolf Nenutil from Masaryk Memorial Cancer Institute for reviewing the results.

The work utilized infrastructure and AI framework developed BioMedAI, supported by European Union’s Horizon Europe research and innovation program under No.~101079183 (BioMedAI TWINNING).
Computational resources were provided by the e-INFRA CZ project (ID:90140), supported by the Ministry of Education, Youth, and Sports of the Czech Republic.

The authors have no relevant financial or non-financial interests to disclose.

\input{paper.bbl}

\end{document}

%% file: paper.bbl

%% file: paper.bbl
\begin{thebibliography}{10}
\providecommand{\url}[1]{#1}
\csname url@samestyle\endcsname
\providecommand{\newblock}{\relax}
\providecommand{\bibinfo}[2]{#2}
\providecommand{\BIBentrySTDinterwordspacing}{\spaceskip=0pt\relax}
\providecommand{\BIBentryALTinterwordstretchfactor}{4}
\providecommand{\BIBentryALTinterwordspacing}{\spaceskip=\fontdimen2\font plus
\BIBentryALTinterwordstretchfactor\fontdimen3\font minus
  \fontdimen4\font\relax}
\providecommand{\BIBforeignlanguage}[2]{{%
\expandafter\ifx\csname l@#1\endcsname\relax
\typeout{** WARNING: IEEEtran.bst: No hyphenation pattern has been}%
\typeout{** loaded for the language `#1'. Using the pattern for}%
\typeout{** the default language instead.}%
\else
\language=\csname l@#1\endcsname
\fi
#2}}
\providecommand{\BIBdecl}{\relax}
\BIBdecl

\bibitem{early-detection}
N.~Mottet, R.~C. {van den Bergh}, E.~Briers \emph{et~al.},
  ``{EAU-EANM-ESTRO-ESUR-SIOG Guidelines on Prostate Cancer—2020 Update. Part
  1: Screening, Diagnosis, and Local Treatment with Curative Intent},''
  \emph{European Urology}, vol.~79, no.~2, pp. 243--262, 2021.

\bibitem{gleason-patterns}
B.~Delahunt, R.~J. Miller, J.~R. Srigley \emph{et~al.},
  ``\BIBforeignlanguage{en}{Gleason grading: past, present and future},''
  \emph{\BIBforeignlanguage{en}{Histopathology}}, vol.~60, no.~1, pp. 75--86,
  Jan. 2012.

\bibitem{gallo}
M.~Gallo, V.~Kraj{\v n}ansk{\'y}, R.~Nenutil \emph{et~al.},
  ``\BIBforeignlanguage{en}{Shedding light on the black box of a neural network
  used to detect prostate cancer in whole slide images by occlusion-based
  explainability},'' \emph{\BIBforeignlanguage{en}{N. Biotechnol.}}, vol.~78,
  pp. 52--67, oct 2023.

\bibitem{CAM}
B.~Zhou, A.~Khosla, A.~Lapedriza \emph{et~al.}, ``Learning deep features for
  discriminative localization,'' in \emph{2016 IEEE Conference on Computer
  Vision and Pattern Recognition (CVPR)}.\hskip 1em plus 0.5em minus
  0.4em\relax Los Alamitos, CA, USA: IEEE Computer Society, jun 2016, pp.
  2921--2929.

\bibitem{GradCAM}
R.~R. Selvaraju, M.~Cogswell, A.~Das \emph{et~al.}, ``{Grad-CAM}: Visual
  explanations from deep networks via gradient-based localization,''
  \emph{International Journal of Computer Vision}, vol. 128, no.~2, p.
  336–359, Oct. 2019.

\bibitem{GradCAMplusplus}
A.~Chattopadhay, A.~Sarkar, P.~Howlader, and V.~N. Balasubramanian,
  ``{Grad-CAM++}: Generalized gradient-based visual explanations for deep
  convolutional networks,'' in \emph{2018 IEEE Winter Conference on
  Applications of Computer Vision (WACV)}.\hskip 1em plus 0.5em minus
  0.4em\relax IEEE, Mar. 2018.

\bibitem{HiResCAM}
\BIBentryALTinterwordspacing
R.~L. Draelos and L.~Carin, ``Use {HiResCAM} instead of {Grad-CAM} for faithful
  explanations of convolutional neural networks,'' 2021. [Online]. Available:
  \url{https://arxiv.org/abs/2011.08891}
\BIBentrySTDinterwordspacing

\bibitem{ROAD}
Y.~Rong, T.~Leemann, V.~Borisov \emph{et~al.}, ``A consistent and efficient
  evaluation strategy for attribution methods,'' in \emph{Proceedings of the
  39th International Conference on Machine Learning}, ser. Proceedings of
  Machine Learning Research, vol. 162, 2022, pp. 18\,770--18\,795.

\bibitem{ROAR}
S.~Hooker, D.~Erhan, P.-J. Kindermans, and B.~Kim, ``A benchmark for
  interpretability methods in deep neural networks,'' in \emph{Advances in
  Neural Information Processing Systems}, vol.~32, 2019.

\bibitem{ehr}
Y.~Zhang, A.~Khakzar, Y.~Li \emph{et~al.}, ``Fine-grained neural network
  explanation by identifying input features with predictive information,'' in
  \emph{Advances in Neural Information Processing Systems}, vol.~34, 2021, pp.
  20\,040--20\,051.

\bibitem{weighting-game}
\BIBentryALTinterwordspacing
L.~Raatikainen and E.~Rahtu, ``The weighting game: Evaluating quality of
  explainability methods,'' 2022. [Online]. Available:
  \url{https://arxiv.org/abs/2208.06175}
\BIBentrySTDinterwordspacing

\bibitem{LRP}
G.~Montavon, A.~Binder, S.~Lapuschkin \emph{et~al.}, \emph{Layer-Wise Relevance
  Propagation: An Overview}.\hskip 1em plus 0.5em minus 0.4em\relax Springer
  International Publishing, 2019, pp. 193--209.

\bibitem{LRPalphabeta}
S.~Bach, A.~Binder, G.~Montavon \emph{et~al.}, ``On pixel-wise explanations for
  non-linear classifier decisions by layer-wise relevance propagation,''
  \emph{PLOS ONE}, vol.~10, no.~7, pp. 1--46, 07 2015.

\bibitem{annotation-agreement}
N.~Wahab, I.~M. Miligy, K.~Dodd \emph{et~al.},
  ``\BIBforeignlanguage{en}{Semantic annotation for computational pathology:
  multidisciplinary experience and best practice recommendations},''
  \emph{\BIBforeignlanguage{en}{J. Pathol. Clin. Res.}}, vol.~8, no.~2, pp.
  116--128, Mar. 2022.

\bibitem{pointing-game}
J.~Zhang, Z.~Lin, J.~Brandt \emph{et~al.}, ``Top-down neural attention by
  excitation backprop,'' in \emph{Computer Vision -- ECCV 2016}, 2016, pp.
  543--559.

\end{thebibliography}
